\documentclass[lettersize,journal]{IEEEtran}
\usepackage{amsmath,amsfonts}
\usepackage{algorithmic}
\usepackage{algorithm}
\usepackage{array}
\usepackage[caption=false,font=footnotesize,labelfont=sf,textfont=sf]{subfig}
\usepackage{textcomp}
\usepackage{stfloats}
\usepackage{url}
\usepackage{verbatim}
\usepackage{graphicx}
\usepackage{cite}

\begin{document}

\title{A VR Training System for Robot-Assisted Vascular Intervention with Dual-Mode Control and Model-Based Interaction Response}




\author{%
Yining Shang,
Shuxiang Guo,
Sheng Cao,
Pengcheng Li,
Bin Wang,
and Mingchao Ding%
\thanks{%
Y. Shang, S. Guo, and S. Cao are with the Key Laboratory of Convergence
Biomedical Engineering System and Healthcare Technology, Ministry of
Industry and Information Technology, Beijing Institute of Technology,
Beijing 100081, China.\par
S. Guo and P. Li are with the Department of Electronic and Electrical
Engineering, Southern University of Science and Technology, Shenzhen,
Guangdong 518055, China.\par
B. Wang and M. Ding are with the Department of Peripheral Vascular
Intervention, Aerospace Center Hospital, School of Life Science, Beijing
Institute of Technology, Beijing 100081, China.\par
{\raggedright
\noindent\hspace*{2em}Corresponding author: Shuxiang Guo\\
\hspace*{2em}e-mail: guo.shuxiang@sustech.edu.cn.\par}%
}%
}

\maketitle

\begin{abstract}
Virtual reality training for robot-assisted vascular intervention requires accurate robotic control and effective feedback on guidewire--vessel interactions. This paper presents a VR training system combining asynchronous dual-mode command mapping with sensorless interaction estimation. The system supports velocity-controlled navigation and position-driven incremental manipulation. Guidewire deformation and a Hunt--Crossley-type contact response are combined and temporally filtered to estimate interaction intensity, while trajectory-based visualization supports post-procedure review. Tests confirmed distinct command-dispatch patterns across three velocity gears and the position mode. During hybrid-scene navigation, the estimated response remained low during straight insertion and increased as the guidewire bent within the vascular model. Trajectory analysis yielded a model-derived cumulative trauma work (CTW) value of 15.166 in normalized units and localized regions of elevated interaction. In a preliminary study with 15 novices, physical-phantom task success increased from 53.3\% to 93.3\% after VR training, while mean completion time decreased from 132.4~s to 96.7~s. These results demonstrate the system's dual-mode operation, deformation-sensitive feedback, and potential to improve subsequent physical-phantom task performance.
\end{abstract}


\section{Introduction}

\IEEEPARstart{V}{ascular} interventional surgery (VIS) enables
minimally invasive treatment of cardiovascular and cerebrovascular
diseases. Robot-assisted VIS (RA-VIS) can improve manipulation
precision and reduce occupational radiation exposure
\cite{ref5,ref6,ref7}. However, navigating flexible guidewires through
tortuous vessels remains demanding because operators must coordinate
instrument motion while interpreting guidewire--vessel interactions.
This creates a need for training systems that reproduce the control
characteristics of robotic platforms and provide interpretable
interaction feedback.

Previous studies have investigated robotic delivery
\cite{ref15}, force- and torque-feedback interfaces \cite{ref16},
virtual-to-real autonomous navigation \cite{ref17}, curvature-aware
delivery control \cite{ref18}, and catheter interaction simulation
\cite{ref9}. Simulation and extended-reality environments have also
been investigated for endovascular training and performance assessment
\cite{ref10,ref11,ref12,ref13}. Although these studies have advanced individual aspects of robot-assisted intervention, few have established an integrated physical-leader--virtual-follower framework that connects robotic manipulation, virtual vascular simulation, interaction feedback, and post-procedure spatial review within one VR training workflow.
In particular, reproducing the visual workflow alone does not ensure
that inputs from a robotic leader-side device are consistently mapped
to virtual guidewire motion or that trainees can interpret the effects
of their manipulation.

Two technical gaps remain. First, operator input, command dispatch,
physics simulation, and visual rendering may operate at different
update rates. Feedback-control and haptically enhanced interfaces have
been studied for intervention teleoperation \cite{ref3,ref8}, but direct
event-by-event forwarding in a hardware-connected simulator may produce
uneven actuation under asynchronous execution. This issue becomes more
complex when continuous velocity commands and discrete position
increments coexist within the same procedure.

Second, interaction feedback in vascular intervention training is
difficult to obtain without dedicated force sensors. Guidewire
deformation and tip--vessel proximity provide complementary cues, but
both are affected by numerical noise, sliding motion, and intermittent
contact. Existing deformable-body simulation, catheter interaction
modeling, and surgical haptic-feedback methods
\cite{ref1,ref2,ref9,ref14} provide useful foundations, but they do not
directly provide a unified sensorless interaction representation for
training. Moreover, instantaneous feedback alone does not record where
interaction occurred along the trajectory or quantify its accumulated
burden for post-training review. Mixed-reality visualization studies
\cite{ref4} motivate such spatially linked feedback.

To address these gaps, this paper presents an integrated
physical-leader--virtual-follower VR training framework for
robot-assisted vascular intervention. The framework reproduces the
manipulation workflow of the robotic platform, provides sensorless
interaction feedback during guidewire navigation, and supports
post-procedure spatial review. Within this framework, a multi-cue
interaction representation method combines distributed guidewire
deformation with local guidewire--vessel proximity to estimate relative
interaction intensity without a force sensor. The main contributions of this paper are:

\begin{itemize}
    \item An integrated physical-leader--virtual-follower VR training
    framework with asynchronous dual-mode command mapping. The framework
    connects physical manipulation, virtual instrument simulation,
    velocity-controlled navigation, position-driven incremental
    manipulation, and real-time interaction feedback within a unified
    training workflow.

    \item A sensorless multi-cue interaction representation method for
    vascular intervention VR training. The method combines distributed
    guidewire deformation with local guidewire--vessel proximity and a
    nonlinear virtual contact response to provide relative
    interaction-intensity feedback without requiring a force sensor.

    \item A trajectory-linked post-procedure analysis function that
    records estimated interaction intensity together with guidewire-tip
    positions and visualizes its spatial distribution along the vascular
    geometry.
\end{itemize}

\section{Teleoperation Platform}

The physical robotic platform adopts a leader--follower teleoperation
architecture. The leader-side console contains two horizontally arranged joystick interfaces for catheter and guidewire manipulation, respectively. These interfaces reproduce the relative
arrangement of the instruments on the operating table and allow the
operator to generate advancement, withdrawal, rotation, and combined
translation--rotation commands. The console can be connected either to
the physical follower robot for robotic intervention or to the computer
running the proposed VR training system. The physical follower robot is
therefore not required during VR training.

The physical follower robot uses four drive units to manipulate as many
as three catheters and one guidewire. Axial instrument motion is produced
by paired friction and pinch rollers. Catheter rotation is transmitted
through a geared Y-valve mechanism, whereas guidewire translation and
rotation are generated by dedicated roller assemblies. This arrangement
supports coaxial catheter--guidewire manipulation and rapid instrument
exchange. Although the simulation platform implements control switching
between the catheter and guidewire, the interaction-estimation method
and quantitative evaluation presented in this study focus primarily on
guidewire--vessel interaction.

\begin{figure*}[!t]
    \centering
    \includegraphics[width=\textwidth]
    {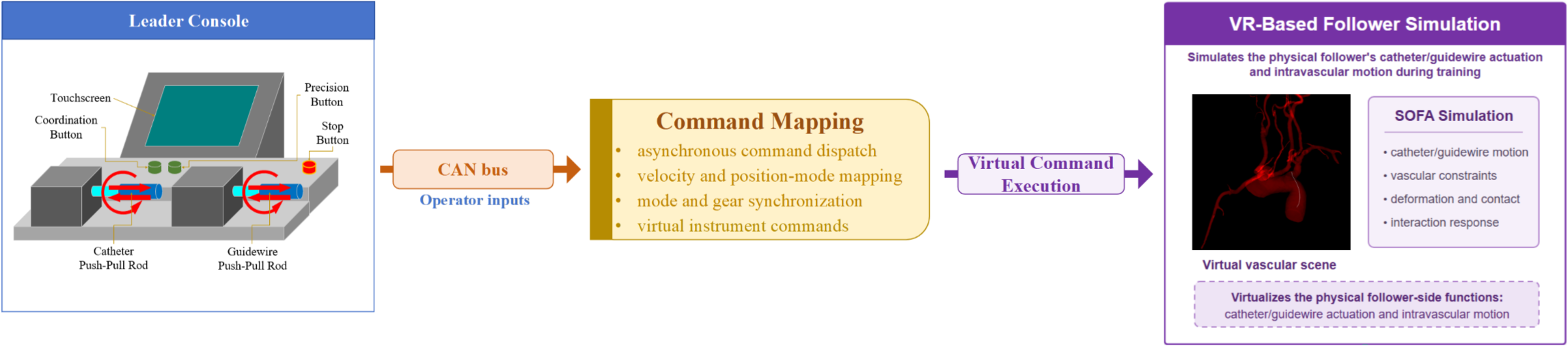}
    \caption{VR training architecture. The physical leader console drives virtual
catheter and guidewire motion in SOFA. The console also supports a physical
follower robot, which is outside the illustrated training loop.}
    \label{fig:tele_platform}
\end{figure*}

The leader-side control interfaces support velocity and position modes.
In velocity mode, the instrument advances or withdraws at one of five
predefined speed levels: 2, 5, 10, 30, and 60~mm/s for Gears~1--5,
respectively. In position mode, the displacement of the control interface
is reproduced as a corresponding displacement of the virtual instrument,
enabling fine adjustment in tortuous branches, near bifurcations, and
within delicate anatomical structures. These control characteristics
form the basis of the velocity and position modes implemented in the VR
training system.

The virtual follower side is implemented using the Simulation Open
Framework Architecture (SOFA), an open-source framework for real-time
interactive mechanical simulation with an emphasis on biomechanics and
medical simulation, rather than the physical follower robot. The catheter and
guidewire are represented as deformable virtual instruments within a
surface-based vascular model. Commands acquired from the leader-side
console are mapped to virtual advancement, withdrawal, and rotation.
During simulation, SOFA updates the instrument configurations under
vascular constraints and provides the guidewire centerline nodes and
instrument-tip positions required for interaction estimation. In the
present implementation, guidewire deformation and tip-to-wall proximity
are used to estimate guidewire--vessel interaction, while catheter motion
is retained as part of the virtual procedural workflow.

Operators must learn to coordinate control modes while interpreting
guidewire bending, intermittent contact, and sliding.

An effective VR system should reproduce these interactions and the
teleoperation control characteristics while maintaining stable command
execution across asynchronous input, simulation, and rendering updates.
The following framework addresses these requirements through dual-mode
command mapping and model-based interaction estimation.

\section{Control and Virtual Interaction Estimation Framework}
\label{sec:control_interaction_framework}

\begin{figure*}[!t]
    \centering
    \includegraphics[width=\textwidth]{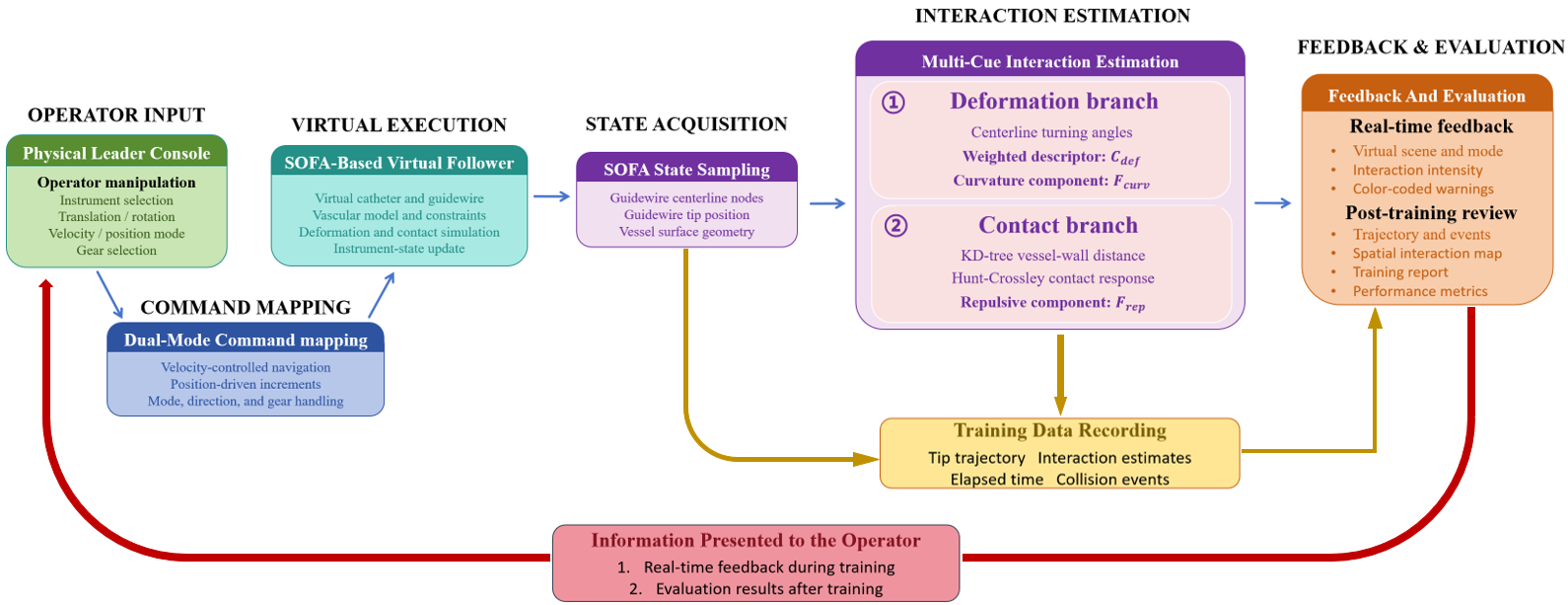}
    \caption{Control and interaction-estimation data flow. Asynchronous commands
drive SOFA instruments; curvature and proximity cues are fused and
filtered for numerical and color-coded feedback. Estimates and tip
positions are recorded for post-procedure analysis.}
    \label{fig:interaction_framework}
\end{figure*}

The proposed framework integrates dual-mode command mapping, SOFA-based virtual instrument execution, state acquisition, interaction estimation, and post-procedure analysis for VR-based vascular intervention training. As illustrated in Fig.~\ref{fig:interaction_framework}, the operator manipulates the catheter or guidewire interface on the leader-side console. The acquired CAN messages specify the selected instrument, motion direction, control mode, and gear level. The asynchronous command-mapping module converts these inputs into executable virtual advancement, withdrawal, or rotation commands. SOFA executes these commands to update catheter and guidewire motion under vascular constraints and contact.

During virtual guidewire navigation, SOFA provides the guidewire centerline nodes and tip position at each simulation step. The centerline nodes are used to calculate the curvature-derived deformation component, while the tip position is used to calculate the proximity-based virtual contact component. The two components are subsequently fused and temporally filtered. The resulting interaction estimate is presented to the operator as a numerical value and a color-coded visual cue on the VR interface and is recorded together with the guidewire-tip position for post-procedure analysis.

\subsection{Asynchronous Dual-Mode Command Mapping}
\label{subsec:command_mapping}

The leader-side console supports velocity and position control modes for
reproducing the manipulation characteristics of the robotic platform.
Velocity mode is used for continuous advancement and withdrawal, with
five gear levels determining the command magnitude and dispatch interval.
Position mode is used for fine manipulation in tortuous vessels and near
bifurcations, where the physical displacement of the leader-side interface
is converted into incremental motion of the virtual instrument.

Because leader-side input events and SOFA simulation updates are
asynchronous, the command-mapping module processes the latest operator
input before dispatching a virtual motion command. In velocity mode,
commands are issued according to the selected direction, gear-dependent
velocity, and dispatch interval. In position mode, small input
fluctuations are suppressed by a displacement deadband, while the remaining
displacement is accumulated and converted into bounded position increments.
This module serves as the control interface between the physical leader
console and the SOFA-based virtual follower.

\subsection{Curvature-Based Deformation Estimation}
\label{subsec:curvature_estimation}

The guidewire deformation state is characterized using directional changes along its discretized centerline. Local bending angles provide a geometric cue for deformation-related interaction and can be computed from the node positions available at each simulation step.

\subsubsection{Guidewire Kinematic Representation}

The guidewire centerline consists of ordered nodes $\{\mathbf{p}_i\}_{i=1}^{N}$, where $\mathbf{p}_i\in\mathbb{R}^3$. Nodes are ordered from the proximal end toward the distal tip. The segment vector and length are defined as
\begin{IEEEeqnarray}{rCl}
    \mathbf{s}_i & = & \mathbf{p}_{i+1}-\mathbf{p}_i,\\
    \ell_i & = & \|\mathbf{s}_i\|_2,
\end{IEEEeqnarray}
where $i=1,\ldots,N-1$. A segment is valid when
$\ell_i>\varepsilon$, with $\varepsilon>0$ denoting a numerical
tolerance. In the implementation, $\varepsilon=10^{-4}$ is used to
exclude degenerate centerline segments. Its normalized direction is
\begin{equation}
    \mathbf{v}_i=\frac{\mathbf{s}_i}{\ell_i}.
\end{equation}
Only pairs of adjacent segments that are both valid are used to calculate
bending angles. The same tolerance is used in all vascular environments
because the guidewire discretization and coordinate scale remain
consistent.

\subsubsection{Robust Curvature Estimation}

For a valid adjacent pair, the bending angle at the shared node $\mathbf{p}_{i+1}$ is
\begin{equation}
    \theta_i=\arccos\!\left(
    \operatorname{clip}(\mathbf{v}_i^\top\mathbf{v}_{i+1},-1,1)
    \right).
\end{equation}
Clipping prevents floating-point errors from placing the dot product outside the domain of the inverse cosine. An angular dead zone suppresses negligible bending and numerical fluctuations:
\begin{equation}
    \tilde{\theta}_i=\max(0,\theta_i-\theta_{\mathrm{th}}),
\end{equation}
where $\theta_{\mathrm{th}}=0.08$~rad suppresses discretization and
simulation noise. This threshold is fixed across vascular environments.

Let $M$ denote the number of valid bending angles, and reindex the retained angles as $\tilde{\theta}_j$, $j=1,\ldots,M$, in proximal-to-distal order. For $M\geq2$, the normalized index and weight are
\begin{IEEEeqnarray}{rCl}
    r_j & = & \frac{j-1}{M-1},\\
    w_j & = & r_j^2.
\end{IEEEeqnarray}
The weighting emphasizes distal bending in the interaction cue. The resulting deformation descriptor is
\begin{equation}
    C_{\mathrm{def}}=\sum_{j=1}^{M}w_j\tilde{\theta}_j.
    \label{eq:deformation_descriptor}
\end{equation}
When $M<2$, the descriptor is set to zero because the normalized weighting is undefined or lacks a valid range.

The scalar $C_{\mathrm{def}}$ summarizes weighted bending along the guidewire. It is a curvature-related geometric descriptor based on turning angles, rather than an arc-length-normalized estimate of local curvature or a direct measurement of contact force.

\subsection{Nonlinear Contact Response and Interaction Fusion}
\label{subsec:interaction_fusion}

The deformation descriptor captures guidewire bending, while tip-to-wall
proximity provides a complementary cue for local virtual contact. The
framework combines these cues through a nonlinear contact-response model
and applies scalar Kalman filtering to reduce temporal fluctuations
associated with numerical noise, sliding motion, and intermittent contact. The curvature, proximity-response, and temporal-filtering stages serve distinct roles in the sequential estimation pipeline and are evaluated here as an integrated method.

\subsubsection{KD-Tree-Based Distance Query}

Let the vascular surface mesh contain $V$ vertices. These vertices are organized in a preconstructed KD-tree $T_{\mathrm{vessel}}$ to support repeated nearest-neighbor queries. At simulation step $k$, the guidewire-tip position $\mathbf{p}_{\mathrm{tip},k}$ is used to obtain
\begin{equation}
    d_k=\mathcal{Q}_{\mathrm{KD}}
    \left(T_{\mathrm{vessel}},\mathbf{p}_{\mathrm{tip},k}\right).
\end{equation}
Each vascular mesh uses the SOFA scene's scale and orientation before
KD-tree construction. The nearest-vertex distance is divided by 100 to
obtain $d_k$, with $d_{\mathrm{safe}}=0.5$ on the same normalized scale.

For a different vascular environment, the tree is rebuilt from its mesh
while normalization and threshold settings remain fixed. Anatomical
differences enter through the queried distance sequence, which
approximates tip-to-surface proximity using mesh vertices.

\subsubsection{Hunt-Crossley-Type Repulsive Response}

A nonlinear virtual contact response incorporates both proximity and approach rate. Let $d_{\mathrm{safe}}$ denote the virtual contact threshold. The effective virtual penetration is
\begin{equation}
    \delta_k=\max(0,d_{\mathrm{safe}}-d_k).
\end{equation}
This quantity represents intrusion into a prescribed proximity zone. When contact is active and a previous distance sample is available, the approach-rate term is approximated as
\begin{equation}
    \dot{\delta}_k\approx\frac{d_{k-1}-d_k}{\Delta t},
\end{equation}
where $\Delta t>0$ is the elapsed time between consecutive simulation samples. The rate term is set to zero when no previous sample is available.

The Hunt-Crossley-type response is
\begin{equation}
    F_{\mathrm{hc},k}=
    \begin{cases}
        \delta_k^n(K+\lambda\dot{\delta}_k), & d_k<d_{\mathrm{safe}},\\
        0, & \text{otherwise},
    \end{cases}
\end{equation}
where the stiffness $K=35$, damping $\lambda=2$, and exponent $n=1.5$
control response magnitude, approach-rate dependence, and nonlinearity,
respectively. These settings are fixed across vascular environments
under the common distance normalization. The repulsive component is
constrained to be nonnegative:
\begin{equation}
    F_{\mathrm{rep},k}=\max(0,F_{\mathrm{hc},k}).
\end{equation}

This constraint prevents an attractive response during separation. For a bounded approach rate, both response terms tend to zero as $\delta_k$ approaches zero, supporting a gradual transition at the virtual contact boundary.

\subsubsection{Interaction Combination and Kalman Temporal Filtering}

The descriptor in Section~\ref{subsec:curvature_estimation} is scaled to
form the curvature-derived component:
\begin{equation}
    F_{\mathrm{curv},k}=k_cC_{\mathrm{def},k},
\end{equation}
where $k_c\geq0$ is the deformation scaling coefficient. In the
implementation, $k_c=10$ is used to scale the weighted curvature
descriptor before it is combined with the virtual contact component.
The two components are expressed on a common interaction-signal scale
and summed as
\begin{equation}
    z_k=F_{\mathrm{curv},k}+F_{\mathrm{rep},k}.
\end{equation}
The resulting signal is a model-derived interaction estimate constructed
from guidewire deformation and tip-to-wall proximity. It is subsequently
filtered and reported in model-derived interaction units for relative
feedback and comparison during VR training.

A scalar random-walk state model and observation model are used for temporal filtering:
\begin{IEEEeqnarray}{rCl}
    x_k & = & x_{k-1}+\eta_k,\\
    z_k & = & x_k+\nu_k,
\end{IEEEeqnarray}
where $x_k$ is the latent interaction state. The process noise $\eta_k\sim\mathcal{N}(0,Q)$ and observation noise $\nu_k\sim\mathcal{N}(0,R)$ are modeled as independent, with $Q\geq0$ and $R>0$.

\begin{figure*}[!t]
    \centering
    \subfloat[VR training setup]{%
        \includegraphics[
            width=0.47\textwidth,
            height=0.25\textwidth,
            keepaspectratio
        ]{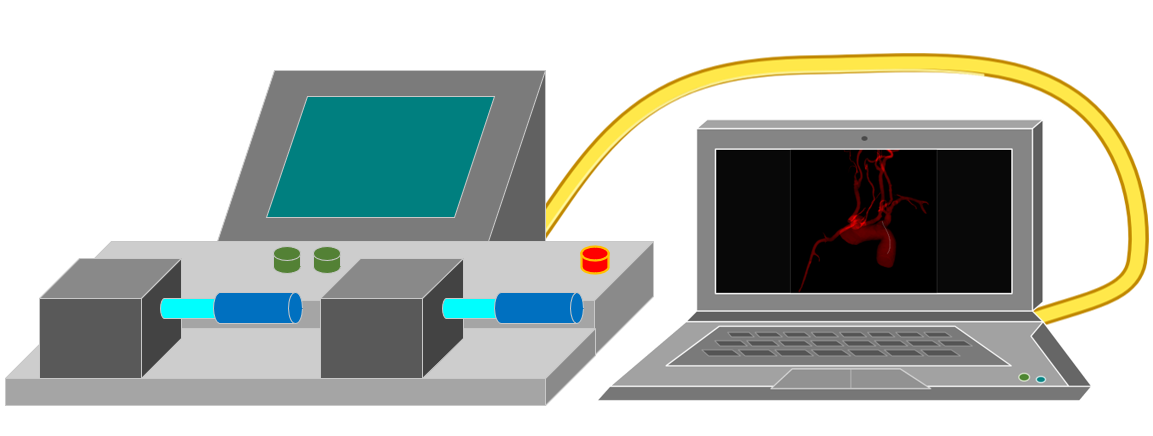}}
    \hfill
    \subfloat[Transparent vascular phantom]{%
        \includegraphics[
            width=0.47\textwidth,
            height=0.25\textwidth,
            keepaspectratio
        ]{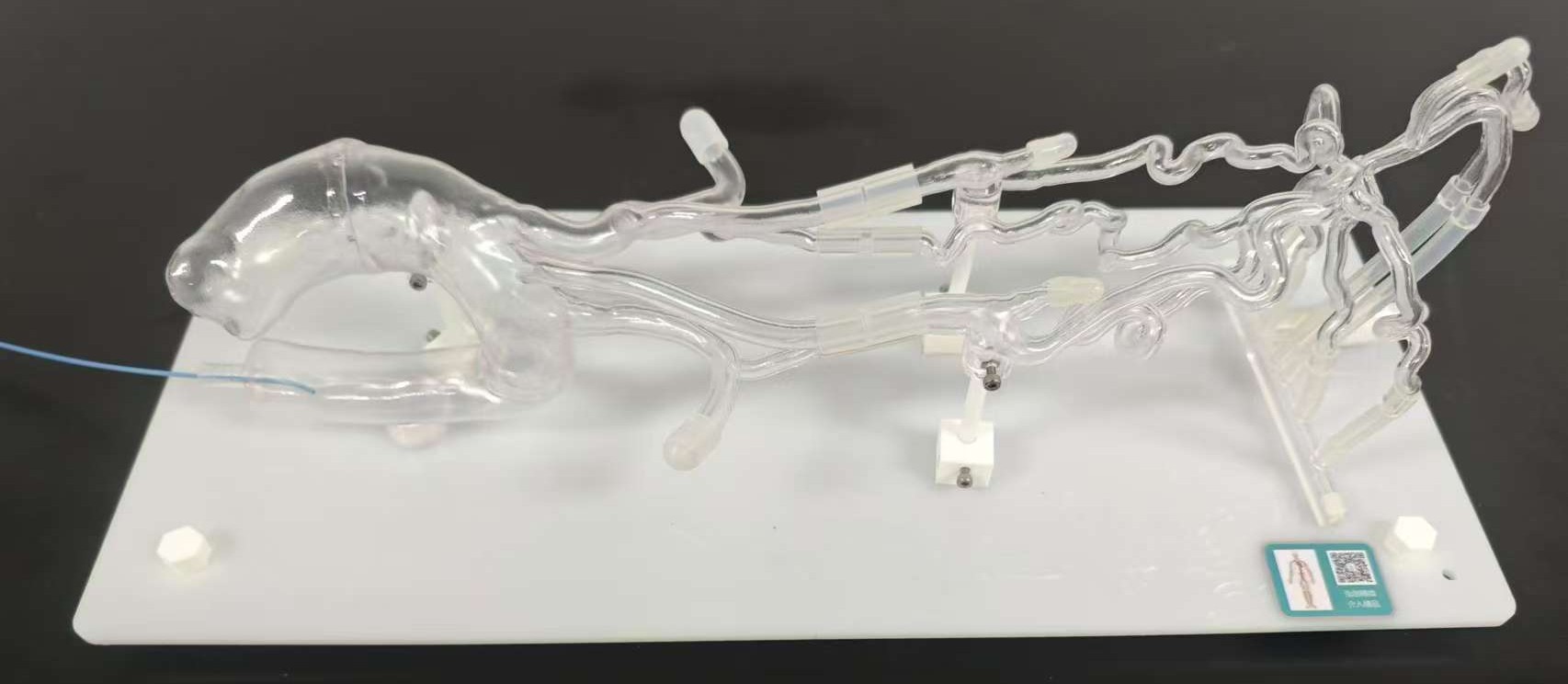}}
    \caption{Experimental components. (a) The joystick serving as the
leader-side control device was connected to a computer running the
SOFA-based vascular intervention training environment. (b) The
corresponding transparent physical vascular phantom was used with the
physical follower robot for physical--virtual comparison and pre--post
training assessment.}
    \label{fig:experimental_setup}
\end{figure*}

The recursive prediction and correction equations are
\begin{IEEEeqnarray}{rCl}
    \hat{x}_{k|k-1} & = & \hat{x}_{k-1},\\
    P_{k|k-1} & = & P_{k-1}+Q,\\
    K_k & = & \frac{P_{k|k-1}}{P_{k|k-1}+R},\\
    \hat{x}_k & = & \hat{x}_{k|k-1}
        +K_k\bigl(z_k-\hat{x}_{k|k-1}\bigr),\\
    P_k & = & (1-K_k)P_{k|k-1}.
    \label{eq:kalman_posterior_covariance}
\end{IEEEeqnarray}
Here, $\hat{x}_{k|k-1}$ and $P_{k|k-1}$ are the predicted state and covariance, $K_k$ is the Kalman gain, and $\hat{x}_k$ and $P_k$ are the posterior estimate and covariance. The contact stiffness $K$ and the time-indexed gain $K_k$ denote different quantities.

The filter parameters are set to $Q=0.05$ and $R=2.0$, with the initial
covariance set to $P_0=1.0$. These settings smooth frame-to-frame
variations in the combined signal while preserving sustained changes caused by guidewire bending and wall approach.

The filter is initialized with $\hat{x}_0=0$. An initialization interval
of $T_{\mathrm{init}}=3.0$~s is used. Whenever the simulation is paused
or the accumulated simulation time $T_k$ is below this interval, the
filtered output is set to zero and the covariance is reset to $P_0$.
After initialization, the filtered estimate $\hat{x}_k$ is displayed to
the operator as a numerical value and a color-coded visual cue and is
recorded for post-procedure analysis. Algorithm~\ref{alg:interaction_pipeline}
gives the complete estimation and filtering procedure, including these
reset conditions.

\begin{algorithm}[H]
    \footnotesize
    \caption{Dynamic Interaction Estimation and Temporal Filtering}
    \label{alg:interaction_pipeline}
    \begin{algorithmic}[1]
        \REQUIRE Nodes $\{\mathbf{p}_i\}_{i=1}^{N}$, tip position $\mathbf{p}_{\mathrm{tip},k}$, $\Delta t>0$
        \REQUIRE Pause state $s_k$, elapsed simulation time $T_k$, $T_{\mathrm{init}}$
        \REQUIRE $T_{\mathrm{vessel}}$, previous distance $d_{k-1}$ if available
        \REQUIRE $K$, $\lambda$, $n$, $k_c$, $d_{\mathrm{safe}}$, $\varepsilon$, $\theta_{\mathrm{th}}$
        \REQUIRE Previous filter state $\hat{x}_{k-1}$, $P_{k-1}$; $Q$, $R$, $P_0$
        \ENSURE $\hat{x}_k$, $P_k$, and distance $d_k$ for the next step
        \STATE Compute $C_{\mathrm{def},k}$ using Section~\ref{subsec:curvature_estimation}
        \STATE $F_{\mathrm{curv},k}\gets k_cC_{\mathrm{def},k}$
        \STATE $d_k\gets\mathcal{Q}_{\mathrm{KD}}(T_{\mathrm{vessel}},\mathbf{p}_{\mathrm{tip},k})$
        \IF{$d_k<d_{\mathrm{safe}}$}
            \STATE $\delta_k\gets d_{\mathrm{safe}}-d_k$
            \IF{a previous distance sample is available}
                \STATE $\dot{\delta}_k\gets(d_{k-1}-d_k)/\Delta t$
            \ELSE
                \STATE $\dot{\delta}_k\gets0$
            \ENDIF
            \STATE $F_{\mathrm{hc},k}\gets\delta_k^n(K+\lambda\dot{\delta}_k)$
            \STATE $F_{\mathrm{rep},k}\gets\max(0,F_{\mathrm{hc},k})$
        \ELSE
            \STATE $F_{\mathrm{rep},k}\gets0$
        \ENDIF
        \STATE $z_k\gets F_{\mathrm{curv},k}+F_{\mathrm{rep},k}$
        \IF{$s_k=\text{paused}$ \OR $T_k<T_{\mathrm{init}}$}
            \STATE $\hat{x}_k\gets0$; $P_k\gets P_0$
        \ELSE
            \STATE $\hat{x}_{k|k-1}\gets\hat{x}_{k-1}$
            \STATE $P_{k|k-1}\gets P_{k-1}+Q$
            \STATE $K_k\gets P_{k|k-1}/(P_{k|k-1}+R)$
            \STATE $\hat{x}_k\gets\hat{x}_{k|k-1}+K_k(z_k-\hat{x}_{k|k-1})$
            \STATE $P_k\gets(1-K_k)P_{k|k-1}$
        \ENDIF
        \STATE Store $d_k$ as the previous distance for the next step
        \RETURN $\hat{x}_k$, $P_k$, $d_k$
    \end{algorithmic}
\end{algorithm}

\subsection{Spatiotemporal Interaction Analysis}
\label{subsec:spatiotemporal_analysis}

Filtered estimates and guidewire-tip positions are recorded for two
complementary analyses: spatial mapping of local interaction intensity
and displacement-weighted accumulation as cumulative trauma work (CTW).

\subsubsection{Cumulative Trauma Work Formulation}

Let $\Delta s_k$ denote the incremental guidewire-tip displacement between consecutive simulation samples:
\begin{equation}
    \Delta s_k=\left\|
    \mathbf{p}_{\mathrm{tip},k}-\mathbf{p}_{\mathrm{tip},k-1}
    \right\|_2.
\end{equation}
Let $\mathcal{I}$ contain the recorded step indices satisfying the minimum-displacement criterion. Cumulative trauma work is defined as
\begin{equation}
    W_{\mathrm{CTW}}=\sum_{k\in\mathcal{I}}\hat{x}_k\Delta s_k.
\end{equation}
This formulation incorporates interaction magnitude and traveled distance, providing a cumulative measure of interaction burden. Unlike an instantaneous peak or a collision count, it accounts for interaction along the recorded path. Because the metric is displacement-weighted, stationary contact does not contribute to the sum.

In this study, CTW is used as a model-derived surrogate metric for
vascular injury risk by combining estimated interaction intensity with
guidewire displacement. By characterizing the interaction accumulated
along the guidewire trajectory, CTW provides a quantitative indicator
for assessing the potential severity and risk of vascular injury. Under
consistent signal-scaling and sampling conditions, higher CTW values
indicate greater accumulated interaction burden and a correspondingly
higher potential for vascular injury.

\subsubsection{Spatiotemporal Heatmap Visualization}

The recorded tuples $\{(\mathbf{p}_{\mathrm{tip},k},\hat{x}_k)\}_{k\in\mathcal{I}}$ are displayed as a color-coded overlay on the vascular geometry. Each tip position is assigned a color according to its associated interaction estimate. Cooler colors indicate lower estimated interaction, and warmer colors indicate higher estimated interaction. A consistent color scale is used when comparing trials.

The overlay relates elevated interaction to bends, bifurcations, and
narrow segments, complementing CTW for post-procedure review.

\section{Experimental Evaluation}
\label{sec:experiments}

The experimental evaluation examined four aspects of the proposed
training system: qualitative agreement between physical and virtual
guidewire configurations, dual-mode command dispatch, the response of
the curvature-derived interaction component to guidewire bending, and
post-procedure visualization of recorded interactions. A preliminary
user study additionally examined whether practice in the VR environment was
associated with improved subsequent performance in the physical vascular
phantom. Together, these experiments characterize both system behavior
and the transfer of training to a physical task.

\subsection{Experimental Setup}

The experimental evaluation used virtual and physical configurations
(Fig.~\ref{fig:experimental_setup}). In the virtual configuration, the
joystick serving as the leader-side control device was connected to a
computer running the SOFA-based VR environment. Commands from the
joystick controlled virtual guidewire and catheter advancement,
withdrawal, and rotation. In the physical configuration, the same
joystick was connected to the physical follower robot, which manipulated
the guidewire and catheter within the transparent vascular phantom. The
physical follower robot was used in the phantom-based experiments but
was not part of the VR training loop.

The vascular model in the VR environment was paired with a corresponding
transparent physical vascular phantom, providing matched vessel paths
for qualitative comparison of guidewire configurations. The
physical--virtual comparison used both configurations, whereas
command-dispatch validation, deformation-related interaction evaluation,
and trajectory-linked visualization were conducted in the VR
environment. The preliminary user study used the physical robotic
configuration for the pre-training and post-training assessments and
the virtual configuration for the intermediate VR training session.
The control pipeline and initialization conditions were kept consistent
across repeated trials of each virtual navigation task.

\subsection{Physical--Virtual Guidewire Configuration Comparison}

\begin{figure}[!htbp]
    \centering
    \subfloat[VR simulation]{%
        \includegraphics[
            width=0.48\columnwidth,
            height=0.35\columnwidth,
            keepaspectratio
        ]{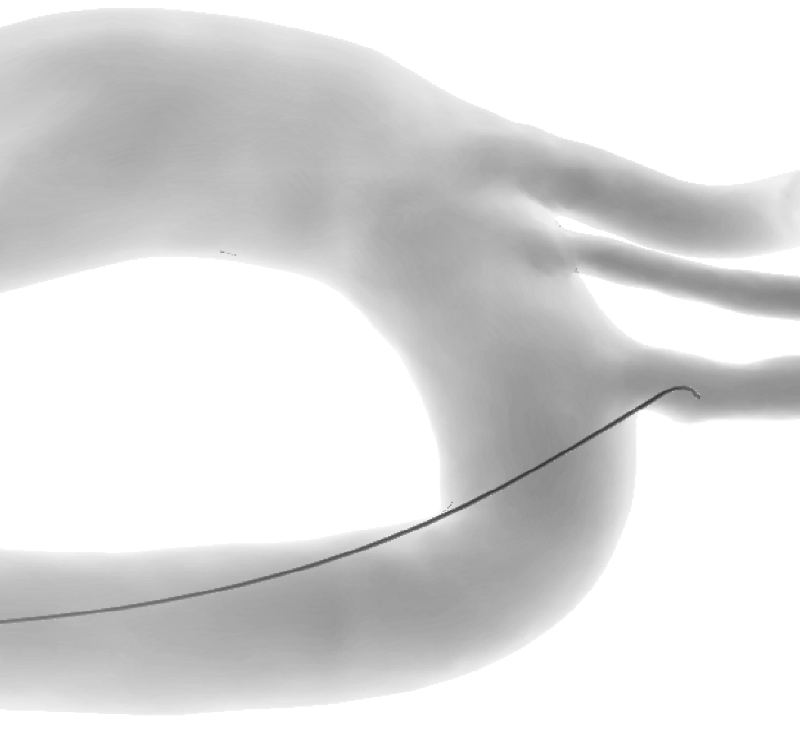}}
    \hspace{0.01\columnwidth}
    \subfloat[Physical vascular phantom]{%
        \includegraphics[
            width=0.48\columnwidth,
            height=0.35\columnwidth,
            keepaspectratio
        ]{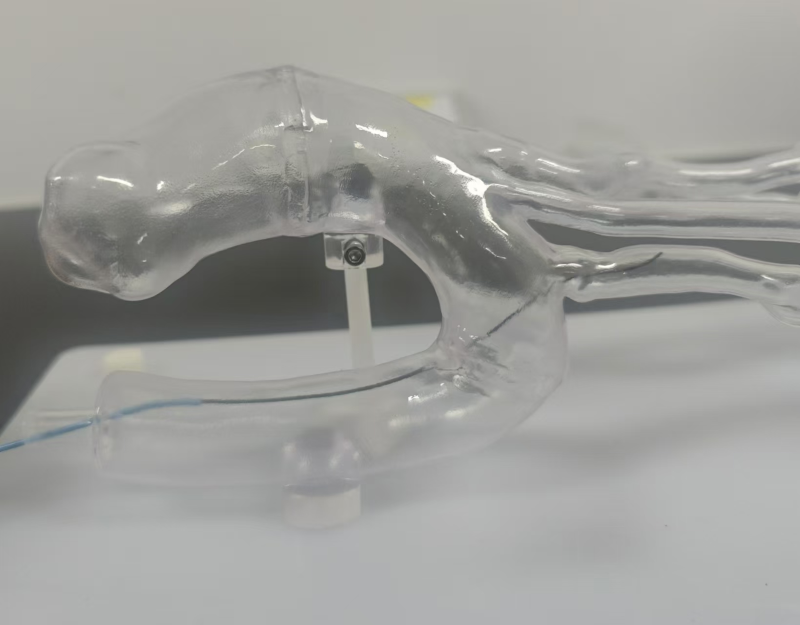}}
    \caption{Qualitative comparison of guidewire configurations in the
    SOFA-based VR environment and the geometry-matched physical vascular
    phantom. The images show guidewire paths through corresponding curved
    vessel segments.}
    \label{fig:physical_virtual_comparison}
\end{figure}

The matched vascular geometries were used to examine whether the virtual
guidewire reproduced the principal deformation pattern visible in the
physical phantom. Fig.~\ref{fig:physical_virtual_comparison} presents
representative configurations in the corresponding vessel segment.

In both images, the guidewire follows the lower curved vessel segment
and bends as it approaches the branch region. The similar large-scale
paths provide qualitative support for the guidewire deformation
represented in the VR environment. This comparison concerns visible
guidewire configuration; the interaction estimates evaluated below are
model-derived signals calculated from virtual deformation and proximity
cues.

\subsection{Dual-mode Control Validation}

The dual-mode controller was evaluated through an equal-distance
traversal test using command-dispatch records from a representative
session. The test examined the stepwise output characteristics of
the velocity gears and position mode. Gear~1, Gear~3, Gear~5, and
position mode each moved the guidewire over the same prescribed
distance, and their activation and dispatch intervals were analyzed.

\begin{figure}[!htbp]
    \centering
    \includegraphics[width=\columnwidth]{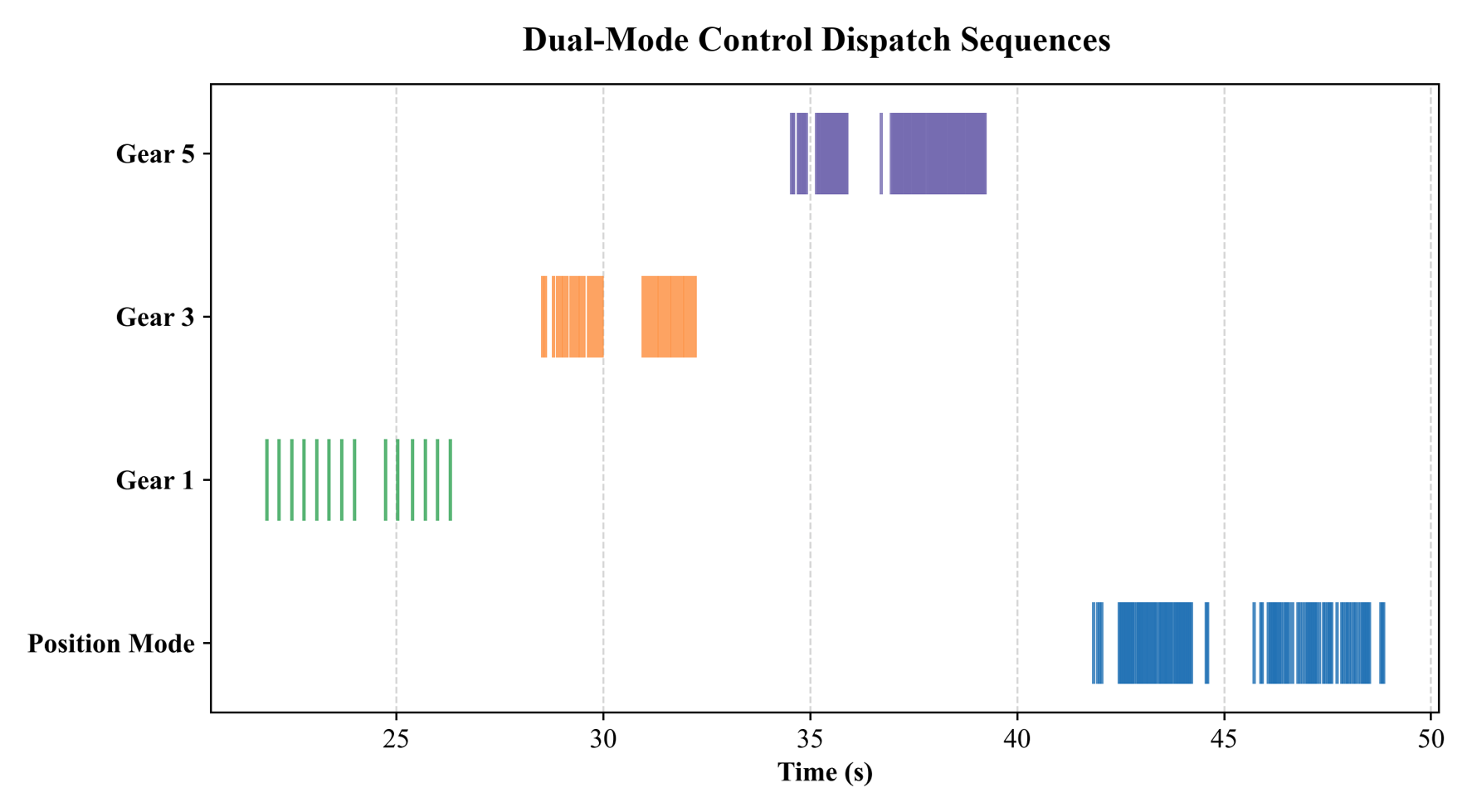}
    \caption{Command dispatch for equal-distance guidewire movements using
Gears~1, 3, and 5 and Position Mode.}
    \label{fig:dual_mode_validation}
\end{figure}

Fig.~\ref{fig:dual_mode_validation} shows distinct dispatch patterns
and traversal durations for equal-distance movements. Lower gears were
used early in the session, followed by higher gears and position
increments, demonstrating mode changes within one workflow.

The record provides a functional demonstration that the different gear
settings and position mode can be used to complete equal-distance
guidewire movements within one training session. 

\subsection{Curvature-Based Interaction Response Evaluation}

A hybrid simulation scene containing a straight glass tube and an
anatomical vascular model was used to examine the response of the
curvature-derived interaction component. The glass tube provided a
low-bending reference segment, while the vascular model introduced curved
paths that constrained guidewire deformation. This configuration allowed
the guidewire to pass from a nearly straight configuration into
progressively bent configurations within the same navigation sequence.

To examine the repeatability of the curvature-derived interaction
response, the navigation experiment was repeated 20 times under the
same simulation settings. Each recorded trajectory was divided into
three representative stages: straight-tube insertion, entry into the
curved vascular segment, and navigation within the deeper curved segment.
For each trial and stage, the curvature-derived response
$F_{\mathrm{curv},k}$ was averaged over all simulation samples within
that stage. The reported values are model-derived responses in arbitrary
units, rather than measured physical forces or maximum response values.

\begin{figure}[!htbp]
    \centering

    \subfloat[]{%
        \includegraphics[width=0.31\columnwidth,trim=0 90 300 70,clip]{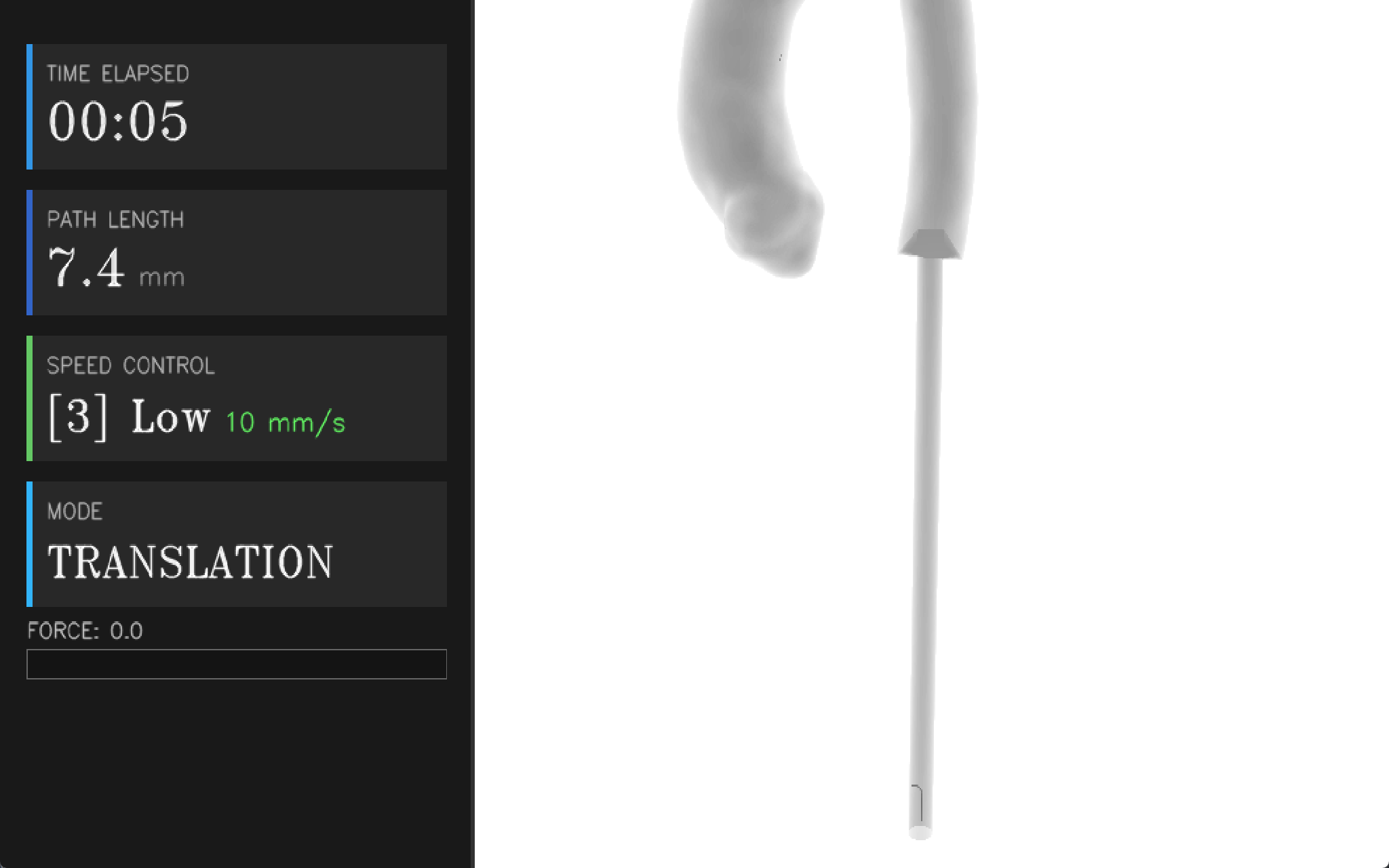}}
    \hfill
    \subfloat[]{%
        \includegraphics[width=0.31\columnwidth,trim=0 90 300 70,clip]{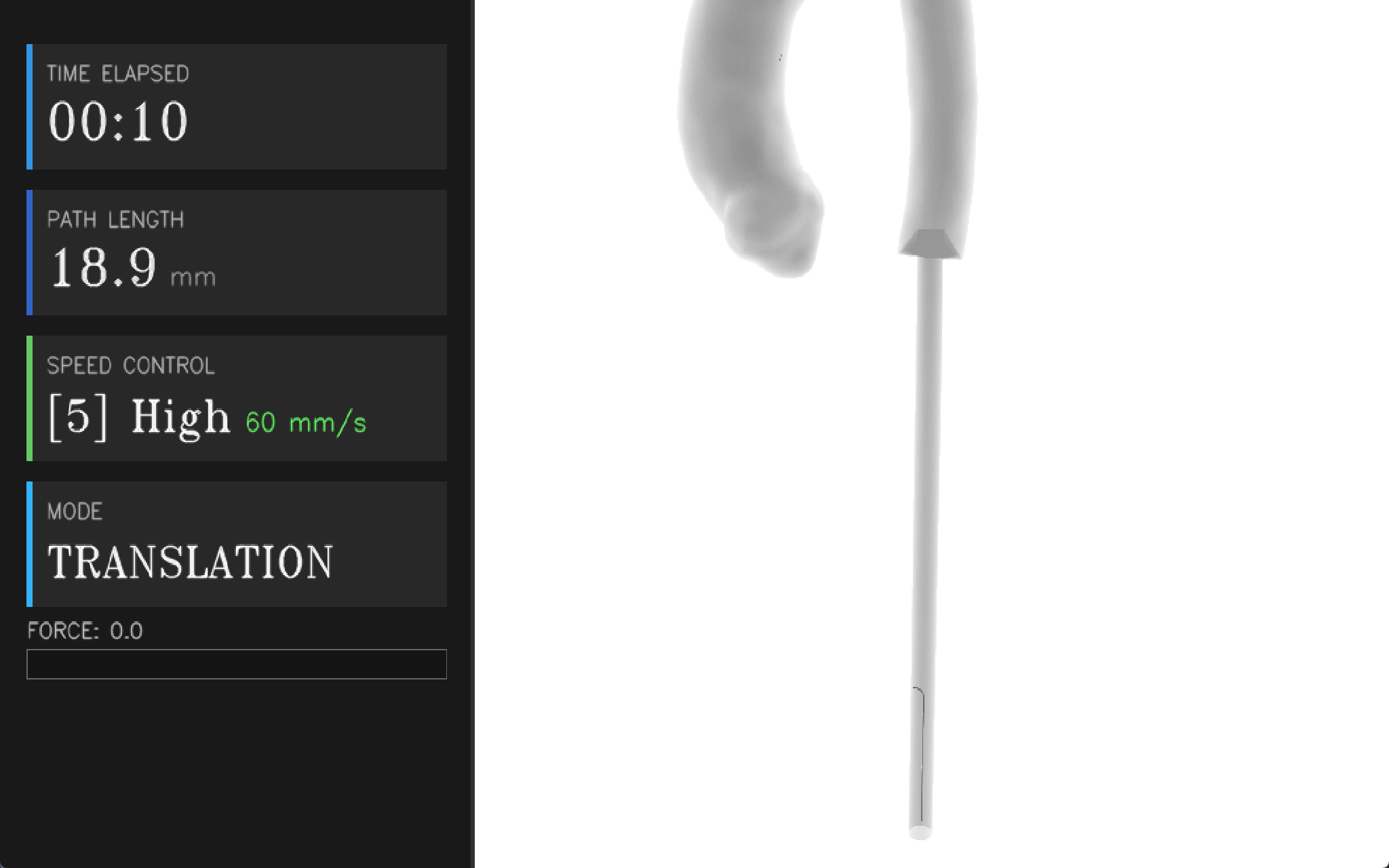}}
    \hfill
    \subfloat[]{%
        \includegraphics[width=0.31\columnwidth,trim=0 90 300 70,clip]{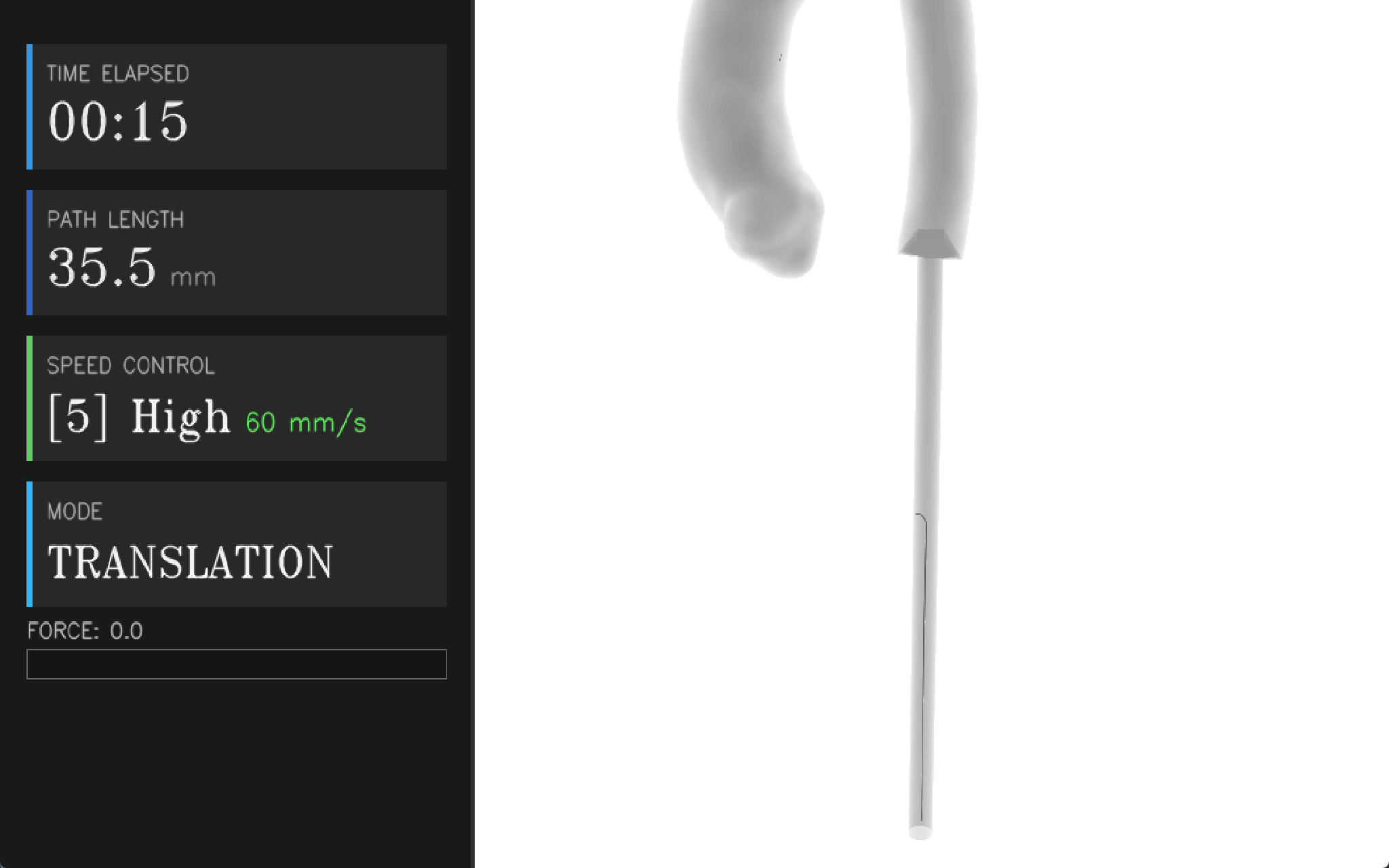}}

    \vspace{-1mm}

    \subfloat[]{%
        \includegraphics[width=0.31\columnwidth,trim=0 90 300 70,clip]{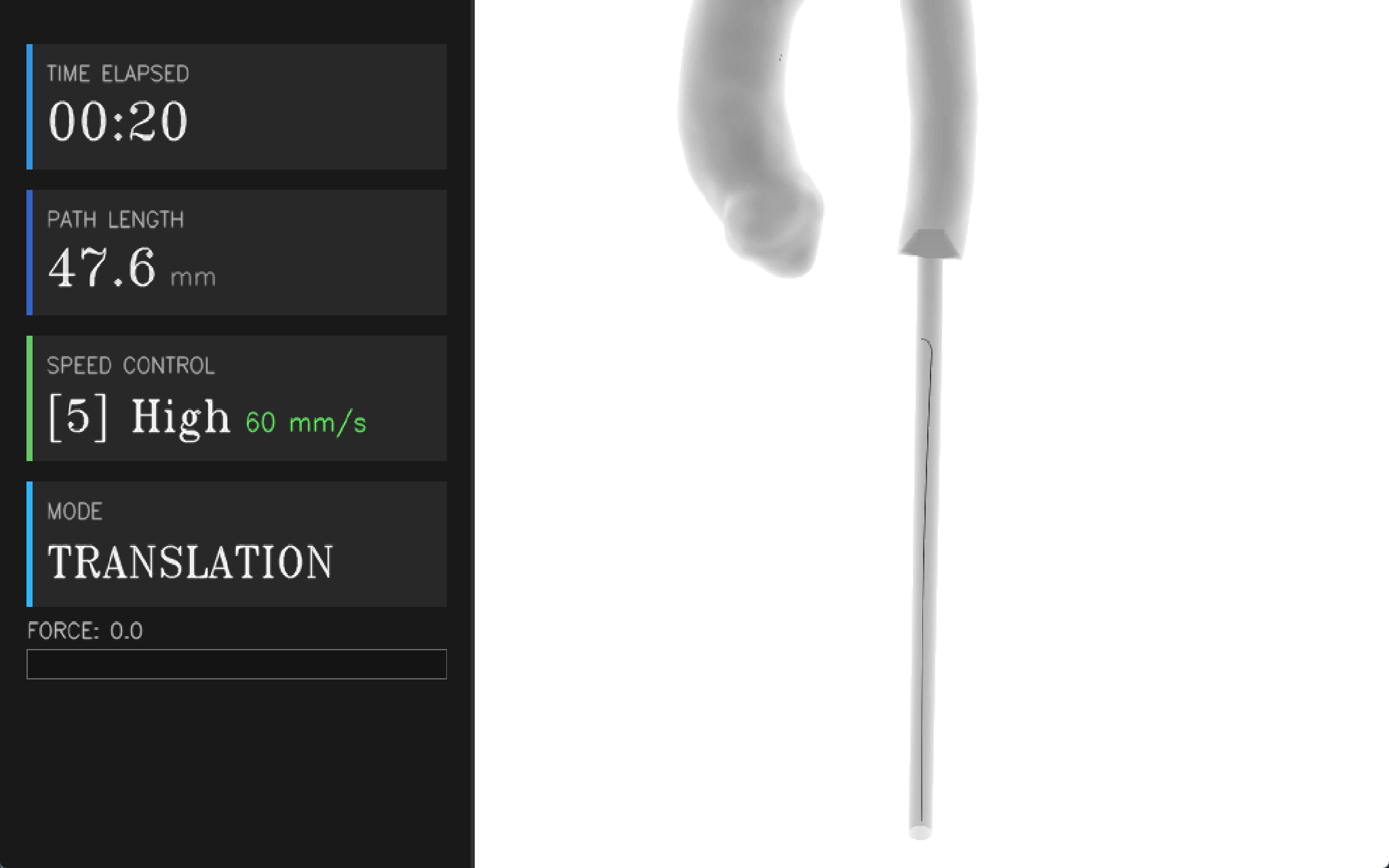}}
    \hfill
    \subfloat[]{%
        \includegraphics[width=0.31\columnwidth,trim=0 90 300 70,clip]{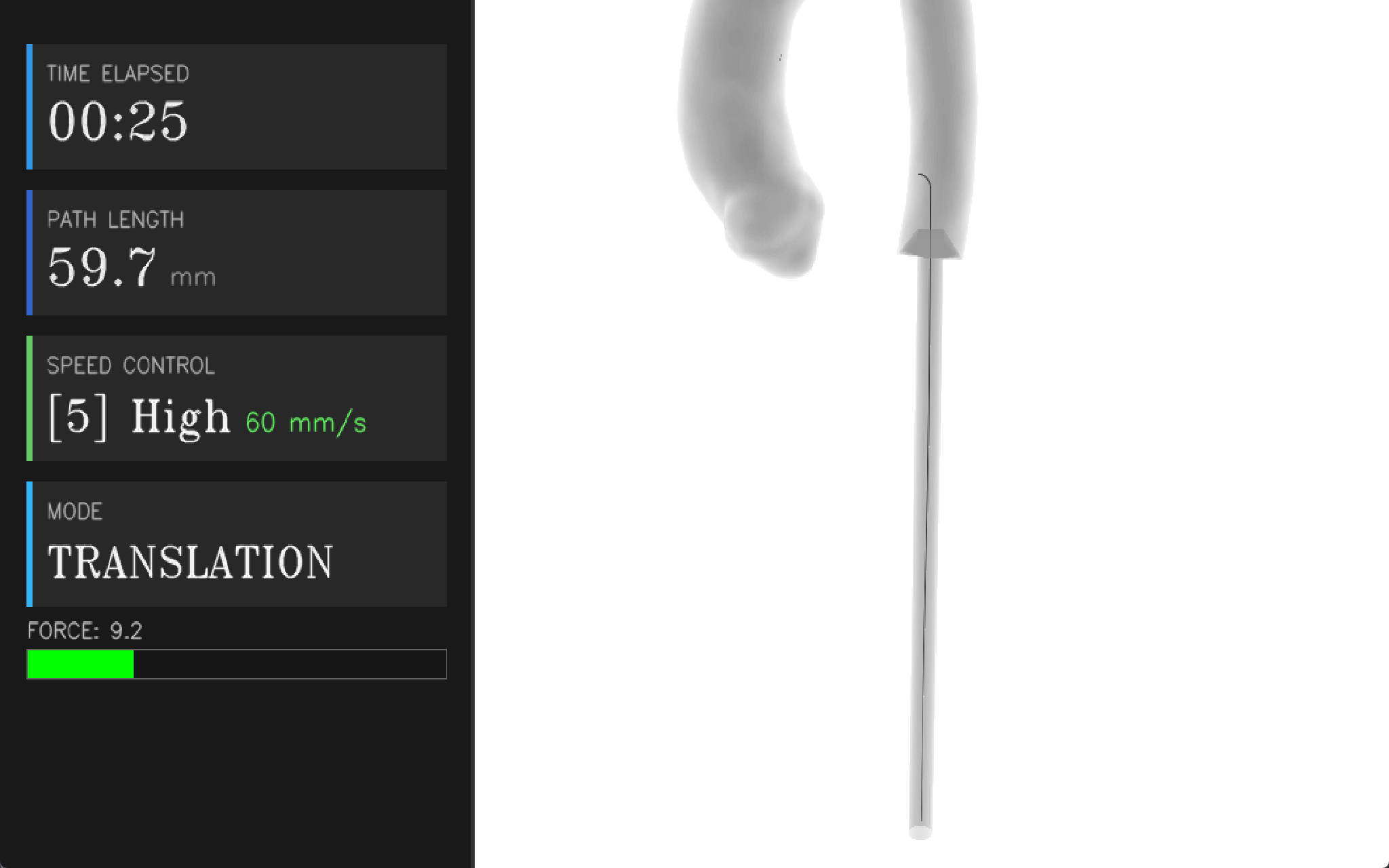}}
    \hfill
    \subfloat[]{%
        \includegraphics[width=0.31\columnwidth,trim=0 90 300 70,clip]{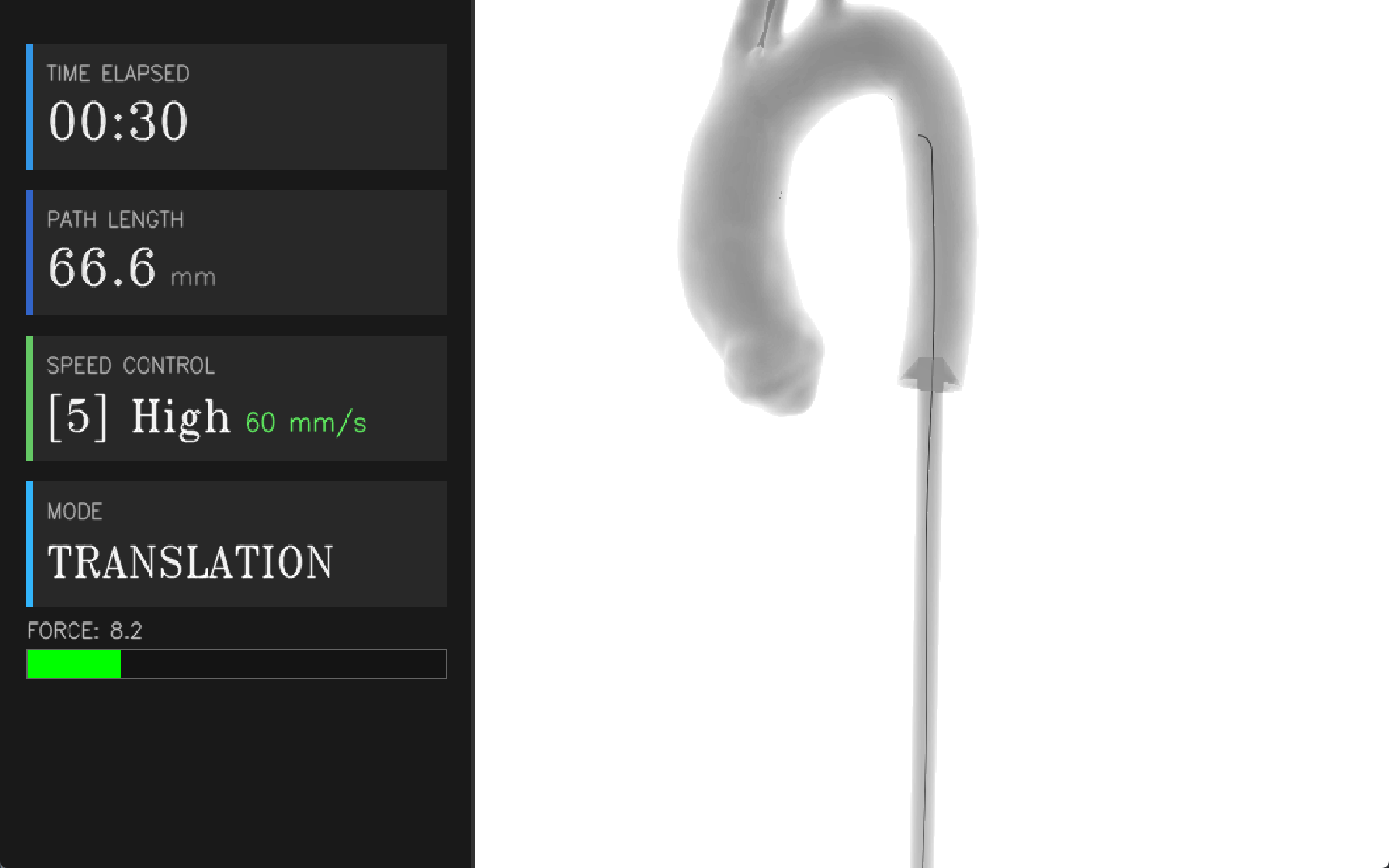}}

    \vspace{-1mm}

    \subfloat[]{%
        \includegraphics[width=0.31\columnwidth,trim=0 90 300 70,clip]{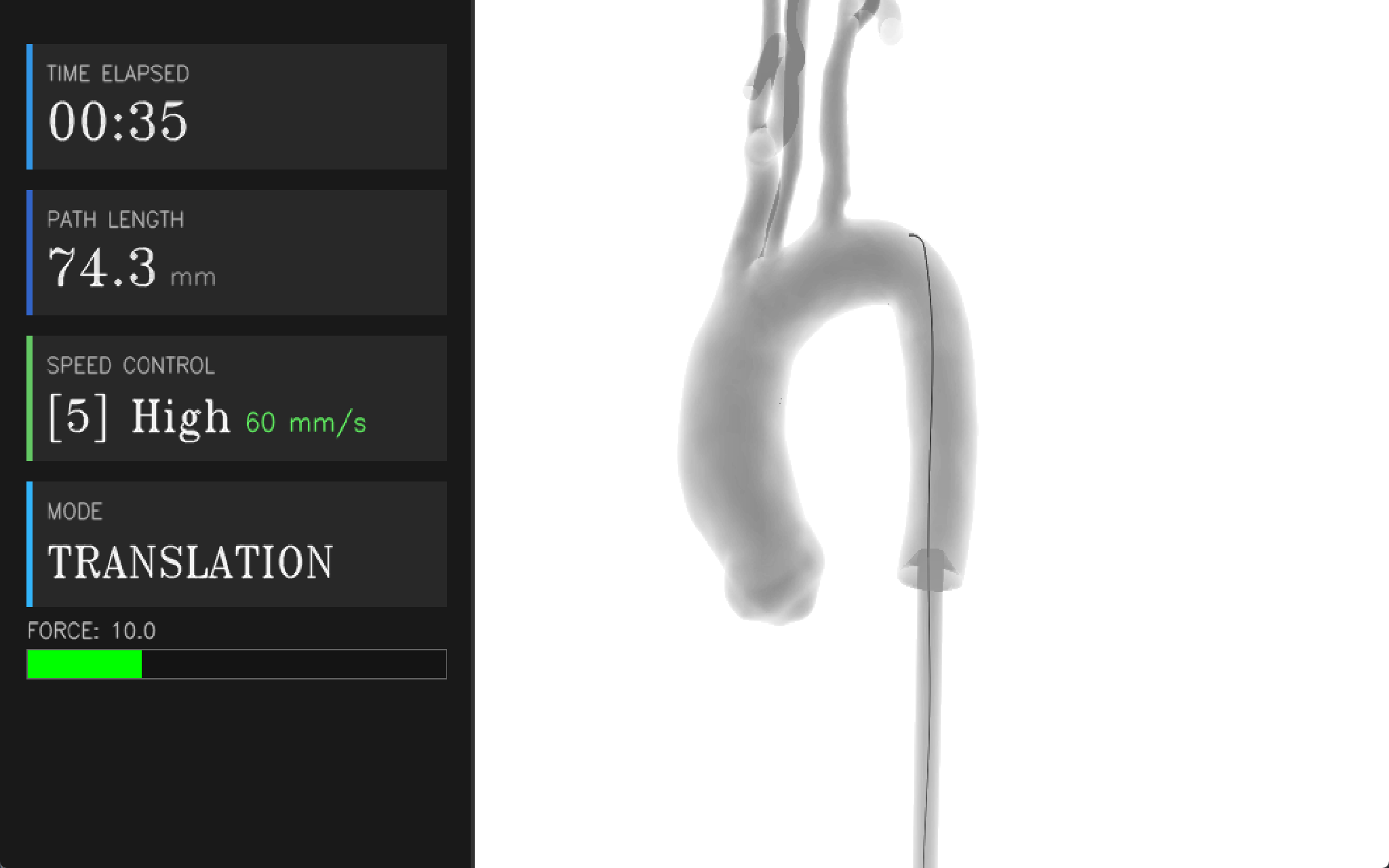}}
    \hfill
    \subfloat[]{%
        \includegraphics[width=0.31\columnwidth,trim=0 90 300 70,clip]{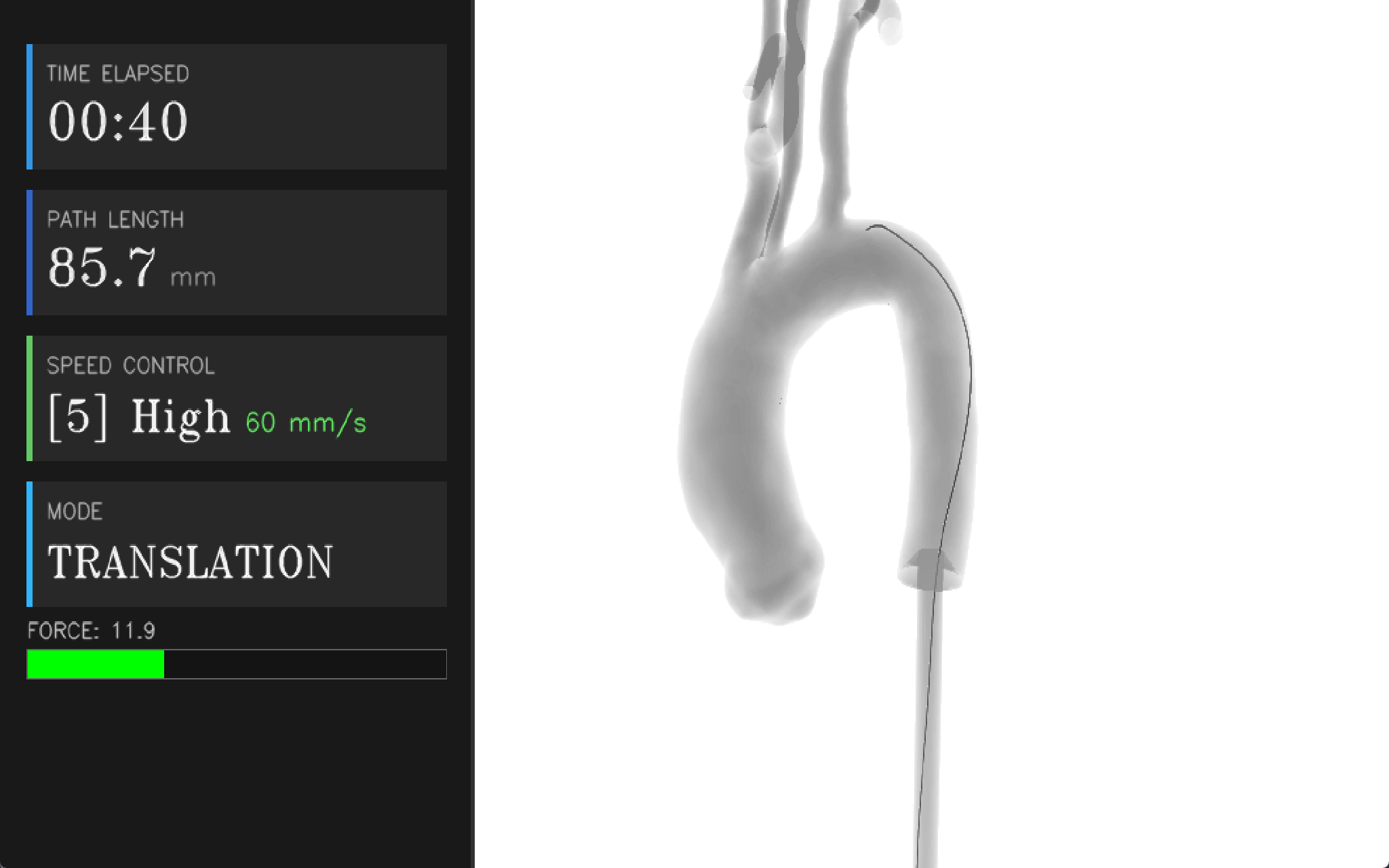}}
    \hfill
    \subfloat[]{%
        \includegraphics[width=0.31\columnwidth,trim=0 90 300 70,clip]{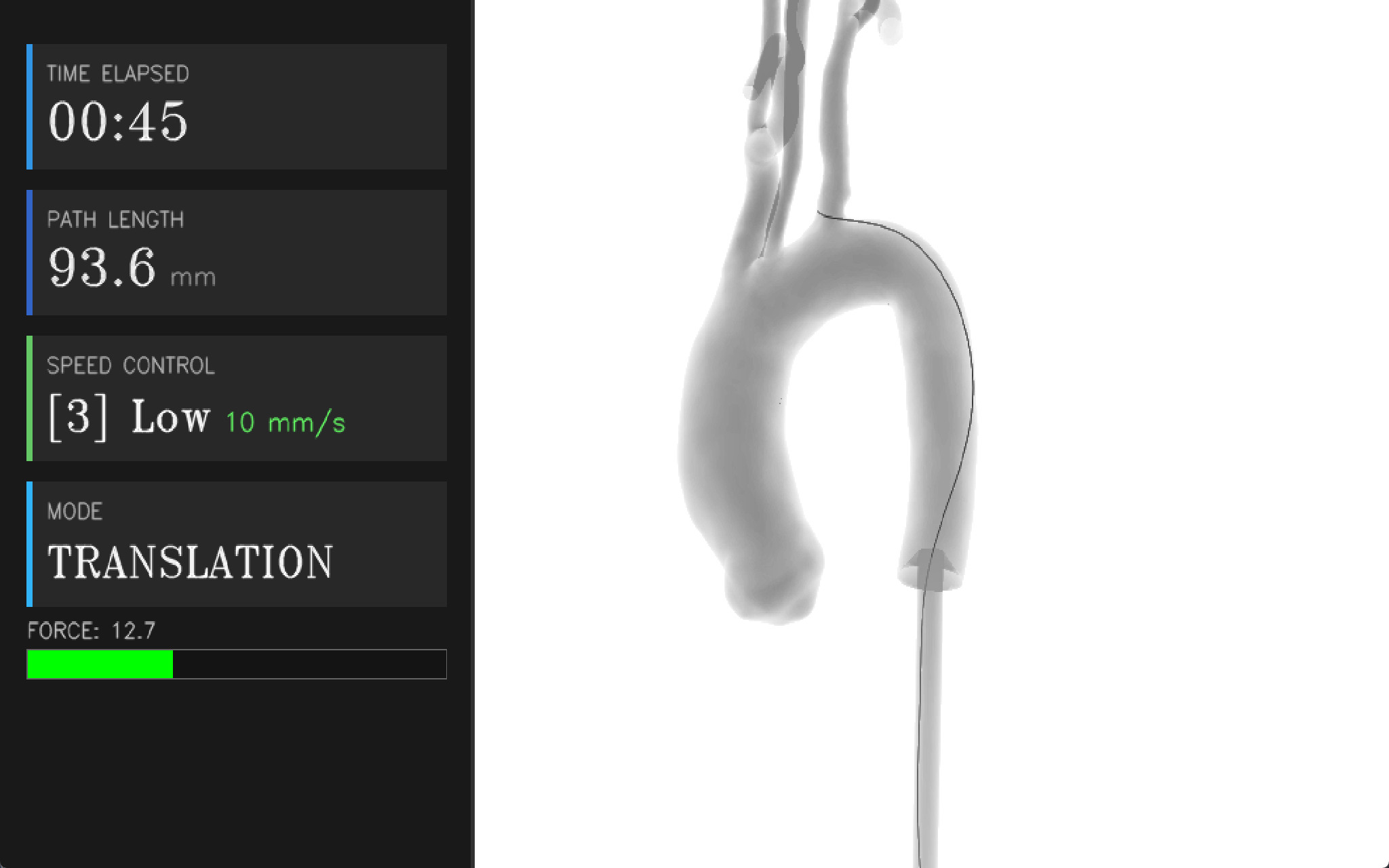}}

    \caption{Sequential snapshots of guidewire navigation and the
curvature-derived interaction response in the hybrid simulation scene.
Panels (a)--(d) correspond to the straight tube (nonvascular segment),
panels (e)--(f) correspond to the curved-vessel entry segment, and
panels (g)--(i) correspond to the deeper curved-vessel segment.
The three stages form consecutive intervals of the same navigation
trajectory and correspond to the stages evaluated in
Table~\ref{tab:curvature_repeatability}.}
    \label{fig:curvature_haptic_grid}
\end{figure}

During advancement through the straight glass tube, the guidewire
centerline remained nearly collinear and the curvature-derived response
stayed close to zero. After entering the vascular model, the guidewire
bent to follow the lumen, and the response increased with the bending
level (Fig.~\ref{fig:curvature_haptic_grid}). Table~\ref{tab:curvature_repeatability}
reports the mean and sample standard deviation of the stage-averaged
responses across the 20 trials, demonstrating an increase from
straight-tube insertion to curved-segment entry and deeper curved
navigation.

\begin{table}[!htbp]
\centering
\caption{Time-averaged curvature-derived interaction responses for
three navigation stages. Values are reported as mean $\pm$ sample SD
across 20 repeated trials.}
\label{tab:curvature_repeatability}

\footnotesize
\setlength{\tabcolsep}{3pt}
\renewcommand{\arraystretch}{1.08}

\begin{tabular}{@{}p{0.43\columnwidth}
                p{0.18\columnwidth}
                p{0.31\columnwidth}@{}}
\hline
\centering Navigation stage &
\centering Expected level &
\centering\arraybackslash Stage-averaged curvature-derived response (a.u.) \\
\hline
Straight tube (nonvascular segment) &
\centering Low &
\centering\arraybackslash $0.0 \pm 0.0$ \\

Curved vessel (entry segment) &
\centering Medium &
\centering\arraybackslash $9.1 \pm 1.5$ \\

Curved vessel (deeper segment) &
\centering High &
\centering\arraybackslash $11.9 \pm 1.4$ \\
\hline
\end{tabular}
\end{table}

\subsection{Spatiotemporal Interaction Visualization}

\begin{figure}[!htbp]
    \centering
    \includegraphics[width=\columnwidth]{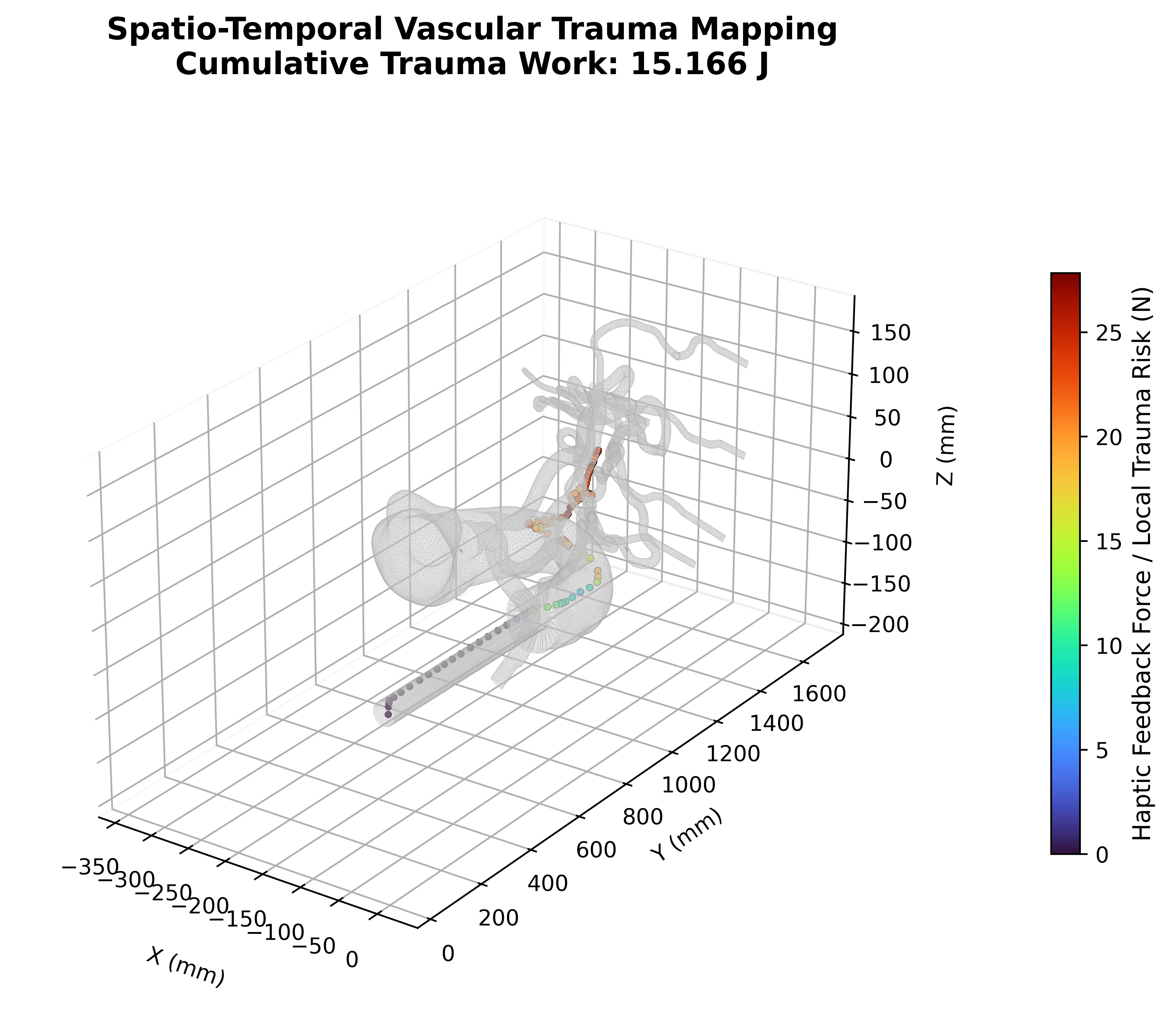}
    \caption{Post-procedure spatial distribution of estimated interaction
    intensity, with CTW summarizing displacement-weighted interaction
    burden.}
    \label{fig:ctw_heatmap}
\end{figure}

After each training session, recorded guidewire-tip positions and
filtered interaction estimates were used to generate a spatial overlay
on the vascular geometry. Samples were color-coded by interaction
magnitude, while cumulative trauma work (CTW) summarized
displacement-weighted interaction burden. The representative heatmap
showed localized regions of elevated interaction separated by
lower-interaction segments (Fig.~\ref{fig:ctw_heatmap}), providing
complementary spatial and cumulative information for post-training
review.

The preceding experiments evaluate the functional behavior of the
control, interaction-estimation, and post-procedure analysis modules.
To examine the training objective of the complete system, a paired
pre-training and post-training study was additionally designed to assess
whether novice users transferred the practiced manipulation skills from
the VR environment to the physical vascular phantom.

\subsection{Preliminary User Training Evaluation}

Fifteen participants without prior experience in robot-assisted vascular
intervention completed a physical-phantom test before and after a
30-min VR training session. Before the initial test, participants
received a standardized introduction to the leader-side console and
task rules without a practice trial. During the physical tests,
participants operated the leader-side joystick, and the follower robot
manipulated the guidewire within the transparent vascular phantom. The
initial guidewire position, target branch, and completion criterion were
identical in the pre-training and post-training tests.

A trial was successful when the guidewire reached the target branch
within 180~s without entering a non-target terminal branch. Completion
time was measured from the first guidewire movement to target acquisition;
unsuccessful trials were assigned the 180-s time limit. Incorrect branch
entries and excessive wall-contact events, defined as visible guidewire
buckling or sustained tip impingement, were independently assessed from
recorded videos by two reviewers, with disagreements resolved by
consensus. Completion time was analyzed using a paired $t$-test after
confirming approximate normality of the within-participant differences.
Paired count and binary outcomes were analyzed using Wilcoxon signed-rank
and McNemar's tests, respectively, with $p<0.05$ indicating statistical
significance. Table~\ref{tab:user_training} summarizes the results.

\begin{table}[!htbp]
\centering
\caption{Physical-phantom task performance before and after VR training
($n=15$). Completion time is presented as mean $\pm$ SD; count outcomes
are presented as median [interquartile range].}
\label{tab:user_training}
\footnotesize
\setlength{\tabcolsep}{2.5pt}
\renewcommand{\arraystretch}{1.15}
\begin{tabular}{@{}lcccc@{}}
\hline
Metric & Pre-training & Post-training & Change & $p$-value \\
\hline
Task success, $n$ (\%)
    & 8 (53.3\%) & 14 (93.3\%) & +40.0 pp & 0.031 \\
Completion time (s)
    & 132.4 $\pm$ 31.8 & 96.7 $\pm$ 24.1 & $-27.0\%$ & $<0.001$ \\
Incorrect entries, $n$
    & 2 [1, 3] & 1 [0, 1] & $-50.0\%$ & 0.008 \\
Excessive wall contacts, $n$
    & 5 [4, 7] & 3 [2, 4] & $-40.0\%$ & 0.004 \\
\hline
\end{tabular}
\end{table}

Task success increased from 53.3\% to 93.3\%, accompanied by a reduction
in mean completion time. The reductions in navigation errors and excessive wall contacts further indicate improved post-training performance. Because the study used a single-group pre--post design, these findings should be interpreted as preliminary evidence rather than as a controlled comparison with conventional training.

\section{Conclusion}

This paper presented a VR training framework for robot-assisted vascular
intervention that integrates asynchronous dual-mode control, model-based
interaction estimation, and post-procedure analysis. The experimental
evaluation verified the operation of the dual-mode controller,
qualitative agreement between physical and virtual guidewire
configurations, the deformation-sensitive interaction response, and the
trajectory-linked review functions.

In the preliminary study with 15 novice participants, physical-phantom
task success increased from 53.3\% before VR training to 93.3\% after
training, accompanied by a reduction in mean completion time. More
importantly, the improved post-training performance in the physical
robotic task suggests that the proposed system can support the transfer
of navigation and manipulation skills practiced in VR to a corresponding
physical intervention environment. The interaction feedback and
trajectory-based review may further help trainees recognize potentially
unsafe manipulation patterns rather than focusing solely on task
completion.

These findings provide preliminary support for the proposed framework
as a training and performance-review tool. Because the participant
evaluation used a single-group pre--post design, however, the observed
improvements cannot be attributed exclusively to VR training. Future
work will validate the interaction estimates against measured forces
and evaluate training effectiveness through randomized controlled
studies with independent training and assessment tasks.


\bibliographystyle{IEEEtran}
\bibliography{references}
\end{document}